\documentclass[pdflatex,sn-mathphys-num]{sn-jnl}

\usepackage{graphicx}%
\usepackage{multirow}%
\usepackage{amsmath,amssymb,amsfonts}%
\usepackage{amsthm}%
\usepackage{mathrsfs}%
\usepackage[title]{appendix}%
\usepackage{xcolor}%
\usepackage{textcomp}%
\usepackage{manyfoot}%
\usepackage{booktabs}%
\usepackage{algorithm}%
\usepackage{algorithmicx}%
\usepackage{algpseudocode}%
\usepackage{listings}%

\theoremstyle{thmstyleone}%
\theoremstyle{thmstyletwo}%

\theoremstyle{thmstylethree}%

\begin{document}

\title[Article Title]{Spontaneous emergence of linguistic statistical laws in images via artificial neural networks}

\author[1]{\fnm{Ping-Rui} \sur{Tsai}}

\author[2]{\fnm{Chi-hsiang} \sur{Wang}}

\author[3]{\fnm{Yu-Cheng} \sur{Liao}}

\author[1]{\fnm{Hong-Yue} \sur{Huang}}%

\author[1*]{\fnm{Tzay-Ming} \sur{Hong}}\email{ming@phys.nthu.edu.tw}


\affil[1]{\orgdiv{Department of Physics}, \orgname{National Tsing Hua University}, \orgaddress{
    \city{Hsinchu} 
    \postcode{30013}, \state{Taiwan}, \country{R.O.C}}}

\affil[2]{\orgdiv{Department of Industrial Engineering and Management}, \orgname{National Yang Ming Chiao Tung University}, \orgaddress{
\city{Hsinchu} \postcode{10587}, \state{Taiwan}, \country{R.O.C}}}

\affil[3]{\orgdiv{Department of Physics}, \orgname{National Taiwan University}, \orgaddress{
\city{Taipei} \postcode{106319}, \state{Taiwan}, \country{R.O.C}}}


\abstract{As a core element of culture, images transform perception into structured representations and undergo evolution like natural languages. Given that visual input accounts for 60\% of human sensory experience, it begs the question of whether images follow similar statistical regularities to linguistic systems. Guided by symbol-grounding theory which posits that meaningful symbols originate from perception, we treat images as vision-centric artifacts and employ pre-trained networks to model the visual processes. By detecting the kernel activations and extracting pixels, we can obtain text-like units which show these image-derived representations adhere to the same statistical Zipf’s, Heaps’, and Benford’s laws as linguistics. Notably, these statistical regularities can spontaneously emerge without explicit symbols or hybrid architectures. This indicates that connectionist networks can automatically develop structured, quasi-symbolic units through perceptual processing alone. It is evident that text- and symbol-like properties can naturally emerge from neural networks, offering a novel perspective for interpretation.}




\maketitle

\section{Introduction}

As Ernst Cassirer noted, ``No longer in a merely physical universe, man lives in a symbolic universe,'' where signs and representations shape our perception of reality. While language constitutes the most explicit manifestation of this symbolic world, other non-linguistic systems, such as images\cite{crosier2007zipf}, music\cite{tsai2024depth}, and genetic sequences\cite{furusawa2003zipf} also encode structured information and exhibit recurring statistical patterns. Among them, Zipf's\cite{saichev2009theory,piantadosi2014zipf}, Heaps'\cite{van2005formal}, and Benford's laws \cite{golbeck2023benford,wang2025novel}respectively describe the scale-invariant relationships in symbol frequency, vocabulary growth, and numerical distributions. The recurrence of these laws across diverse domains suggests that they are not unique to language, but reflect general principles that govern all symbolic organization.

Although these statistical laws have been widely observed in natural languages and other symbolic systems, their origins remain debated. Classical explanations attribute them to communicative optimization, cognitive constraints, or the principle of least effort\cite{zipf1949principle}; however, these accounts typically presuppose the existence of discrete symbols and explicit semantic structures. This raises a central question: can language-like statistical organization spontaneously emerge in systems that are neither explicitly symbolic nor designed for linguistic processing? Related to the symbol grounding problem in cognitive science, this question is crucial to understanding how symbolic structures arise from sub-symbolic representations. In this context, visual perception plays a pivotal role since the origins of human symbolic systems are often closely linked to cognitive capacities for recognizing images, shapes, and spatial features. Understanding how visual features are decomposed and represented is, therefore,  key to studying the emergence of symbols\cite{harnad1990symbol}.

The fundamental elements of images are concerned with how we decompose and interpret visual information. The extraction of image features can be traced back to the pioneering experiments of David Hubel and Torsten Wiesel on the visual cortex of cats, for which they were awarded the 1981 Nobel Prize in Physiology or Medicine\cite{hubel1963visual,hubel1998early}. Their discovery of orientation-selective cells in the primary visual cortex laid the groundwork for understanding the hierarchical nature of visual processing. Inspired by these findings, early computational models, such as Neocognitron\cite{fukushima1983neocognitron}, were developed to replicate biological mechanisms of pattern recognition. Furthermore, statistical properties of luminance distributions, often described in terms of order parameters\cite{reynaud2019second,emrith2010measuring}, play a central role in determining how visual stimuli are perceived and categorized. In particular, the spectral composition of an image, revealed through two-dimensional Fourier analysis, shows that variations in 
high- and low-frequency components significantly influence scene and object categorization tasks\cite{torralba2003statistics}. These surface texture features provide a natural basis for analyzing and structurally representing images.


Deep visual neural networks provide an ideal experimental platform for investigating this question, particularly in light of recent advances in brain–computer interfaces \cite{higgins2021unsupervised,du2022fmri,han2019variational}and studies demonstrating that pre-trained convolutional neural networks (Pre-CNNs)\cite{eickenberg2017seeing} optimized for human-level recognition and multi-label perception exhibit strong correspondences with human visual information processing. Although these models are purely connectionist systems\cite{harnad1990symbol} operating on continuous representations,lacking explicit symbolic manipulation or linguistic supervision, evidence from both neuroscience and artificial intelligence suggests that the hierarchical feature representations learned by deep visual networks closely mirror the stages of human perceptual processing\cite{eickenberg2017seeing}. 
This convergence raises the possibility that the internal feature maps of trained networks may spontaneously organize into structured units that resemble symbolic systems, even in the absence of predefined symbols or semantic constraints.

Understanding whether and how symbolic structure can emerge from purely perceptual representations remains a central question in cognitive science, neuroscience, and artificial intelligence\cite{eickenberg2017seeing,steels2008symbol}. Pre-CNNs, which operate as connectionist systems on continuous visual inputs, offer a natural testbed for investigating this issue. In particular, their hierarchical feature representations have been shown to closely align with stages of human visual processing, motivating the question of whether visual features may exhibit organizational properties analogous to those observed in language.

From the perspective of statistical linguistics, written language does not derive its structure from isolated symbols, but from statistical regularities that arise through interactions among elements\cite{hogan2021knowledge}, such as the interpretation of the linguist John Rupert "You shall know a word by the company it keeps”\cite{maiden1992irregularity}. Despite their apparent complexity, natural languages exhibit robust and reproducible scaling laws. A prominent example is Zipf's law~\cite{piantadosi2014zipf}, which describes a power-law relationship between word frequency $P(x)$ and its rank $x$, $P(x) \sim x^{-\alpha}$. Similar statistical patterns have been identified beyond language, including music~\cite{tsai2024depth,zanette2006zipf}, genomic sequencing\cite{furusawa2003zipf} and painting~\cite{crosier2007zipf}. This suggests that these properties are likely not unique to linguistic symbols, but general to all structured representation.
Two additional regularities in linguistics are Heaps' law~\cite{gelbukh2001zipf}, which characterizes the growth of vocabulary size as a function of text length, and Benford's law~\cite{miller2015benford}, which governs the distribution of leading digits in numerical data and has recently been shown to extend to written texts across multiple languages~\cite{golbeck2023benford}. Together, these three laws provide a statistical lens through which structured organization can be examined independently of their semantic interpretation.

Motivated by these observations, Motivated by these observations, this work investigates whether language-like statistical regularities arise in the visual representations learned by Pre-CNNs. To this end, we introduce an analysis framework that defines visual ``words'' based on the activation patterns of individual convolutional kernels. The frequency of each visual word is quantified by counting pixels whose activation exceeds a fixed proportion of the maximum response within a feature map. This definition allows language-inspired statistical analyses to be applied directly to visual data without imposing explicit symbolic labels or semantic supervision. Using this framework, we systematically evaluate whether Zipf's, Heaps', and Benford's laws emerge across different layers and architectures of Pre-CNNs.

This study addresses three main objectives: (1) define the equivalent of words in an image to enable statistical analysis in Sec. 2.1; (2) examine whether language-like statistical scaling laws emerge across layers and architectures of pre-trained CNNs in Sec. 2.2; and (3) test the robustness of these statistical laws under adversarial perturbations and corrupted inputs in Sec. 2.3.

\section{Results}\label{sec2}
\subsection{What plays the role of words in an image?}

When applying the skills in statistical linguistics to image analysis, the first essential step is to define what constitutes ``words'' within an image. This is achieved by appealing to the convolutional kernels in pre-trained convolutional neural networks. Each kernel consists of Gabor-like orientation-selective filters that extract edge and texture features\cite{zeiler2014visualizing}, functionally analogous to the receptive fields of simple cells\cite{carandini2006simple} in the primary visual cortex. Within this framework, different convolutional kernels are treated as distinct morpheme types\cite{carlisle2005exploring}. The feature maps produced by convolving these kernels with the input from the preceding layer encode the spatial locations and activation strengths of the corresponding morphemes within an image. 

To quantify the occurrence of each morpheme, we apply a thresholding procedure to each feature map, selecting pixels whose activation values exceed 90\% of the maximum activation in that map\cite{selvaraju2017gradcam,zhou2016learning,van2019deep}. This approach retains the most salient response regions and follows a strategy commonly used in deep learning to identify dominant feature activations. The number of such highly activated pixels is taken as its occurrence frequency. By ranking these morpheme frequencies in descending order, we can obtain distributions to compare with that of Zipf’s law. When each convolutional kernel is treated as a unit and the cumulative number of word tokens and types is counted sequentially, Heaps’ law can also be derived. Finally, to assess the Benford's law, we group feature maps across different convolutional layers into nine hierarchical sets and analyzing the resulting word-frequency distributions. Please refer to Secs. 5.1$\sim$3 in  Methods for detailed settings and procedures,.

\begin{figure}
\centering
\includegraphics[width=13cm]{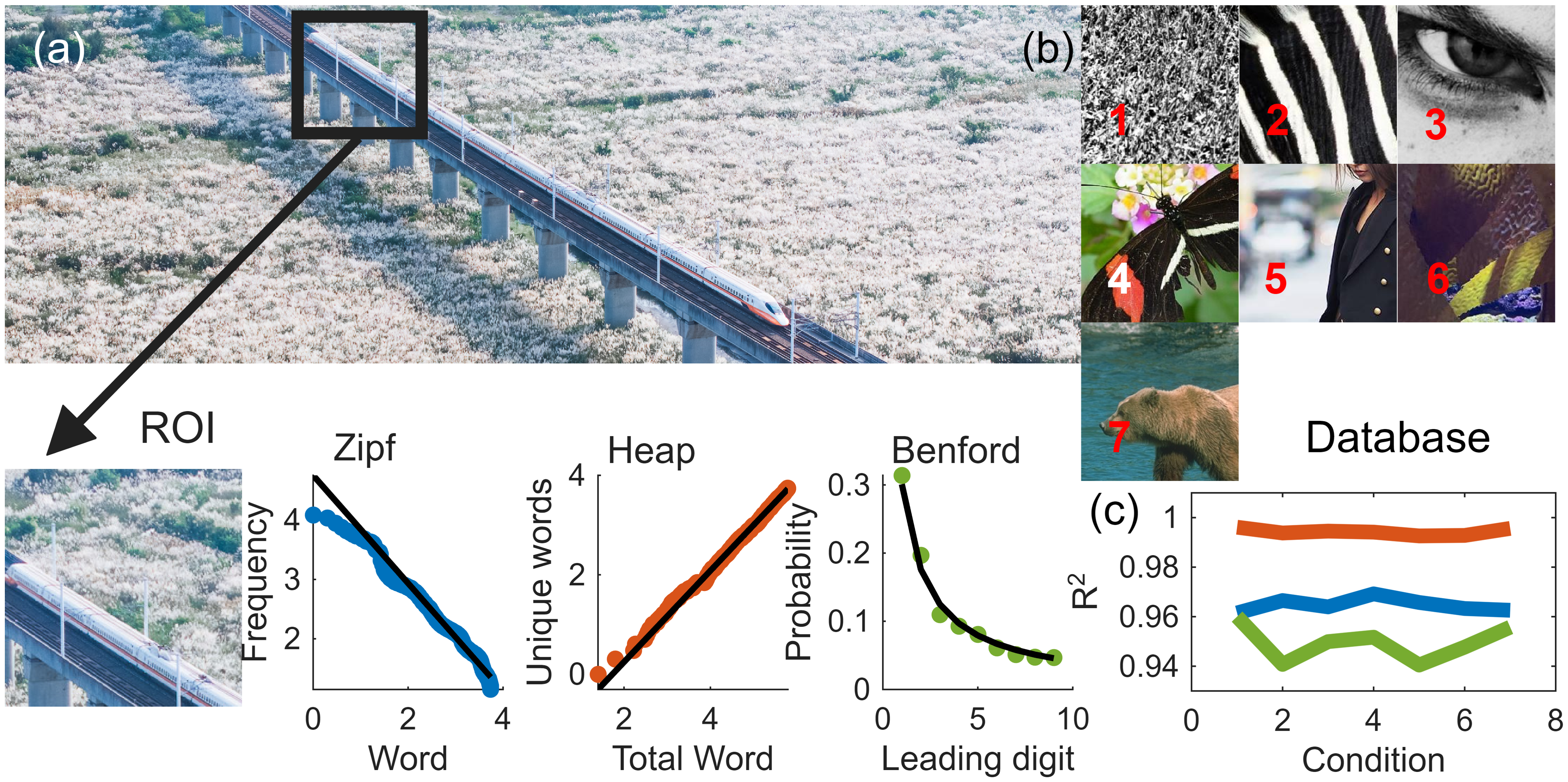}\caption{\textbf{Three laws in statistical linguistics emerging in images and databases}. 
In (a), we used a landscape of Taiwan photographed by Wei-Hsiung Huang (foto WH) and extracted a 224$\times$224 RGB Region of Interest (ROI). This ROI was then fed into the pre-trained CNN VGG-19, resulting in the emergence of Zipf’s, Heaps’, and Benford’s laws, shown respectively in blue, orange, and green. (b) illustrates the surface-texture characteristics of seven image databases, which we define as the experimental conditions. 
(c) shows the R-squared results by inputting 16 images from each of the seven conditions into our nine pre-trained CNNs. 
The color scheme is the same as in (a). 
R-squared values above 0.93 suggest that the regression lines represent the data well.}
\end{figure}

In linguistics, meaning is often understood as emerging from relational structure rather than intrinsic properties of isolated symbols. Words acquire meaning through their patterns of co-occurrence and mutual constraints within a network of relations, an idea formalized in distributional semantics, structural linguistics, and widely adopted in knowledge graphs and symbolic systems. Within this perspective, semantic content is not assigned a priori, but arises from contextual dependence across a structured system. Pre-CNNs constitute a fundamentally connectionist form of visual processing that operates entirely on continuous activations and local interactions. Crucially, such models are not endowed with any linguistic symbols, semantic labels, or conceptual priors. As a result, they do not presuppose the existence of symbolic meaning and therefore avoid the circularity inherent in the symbol-grounding problem\cite{harnad1990symbol}, often referred to as the ``symbol grounding carousel.'' Any structured, language-like regularities observed in these networks must instead arise endogenously from perceptual organization and task-driven learning dynamics, rather than predefined symbolic representations.

In Fig. 1(a), we selected a landscape photograph of Taiwan for analysis. Because each Pre-CNN has constraints on the input image resolution, we extracted a Region of Interest (ROI) for processing. After applying a bilateral log10 transformation to the data, we obtained the blue/orange/green distributions corresponding to Zipf’s/Heaps’/Benford’s laws.
To examine how variations in surface texture influence the statistical behavior associated with the three statistical laws, we employed seven publicly available image databases as experimental conditions, selecting sixteen images from each dataset for analysis. Representative examples and visual characteristics of these image sets are shown in Fig.~1(b), while the corresponding data licenses are detailed in Method Sec.~4.3. Each image set was processed using nine distinct Pre-CNNs: VGG16 (VG16) and VGG19 (VG19)~\cite{simonyan2014vgg}, DarkNet-19 (D19) and DarkNet-53 (D53)~\cite{redmon2016darknet}, EfficientNet-b0 (EF0)~\cite{tan2019efficientnet}, Inception-v3 (INV3)~\cite{szegedy2016rethinking}, DenseNet-201 (D201)~\cite{huang2017densenet}, MobileNet-v2 (MOBV2)~\cite{sandler2018mobilenetv2}, and ResNet-18 (RE18)~\cite{he2016resnet}. The goodness-of-fit results, quantified using the coefficient of determination ($R^2$), are summarized in Fig.~1(c). Notably, across all datasets and network architectures, the fitted distributions consistently achieved $R^2$ values exceeding 0.93, indicating a robust adherence to the corresponding statistical laws. In the following, we use P to denote probability and UW to denote unique words.



\begin{figure}
\centering
\includegraphics[width=13cm]{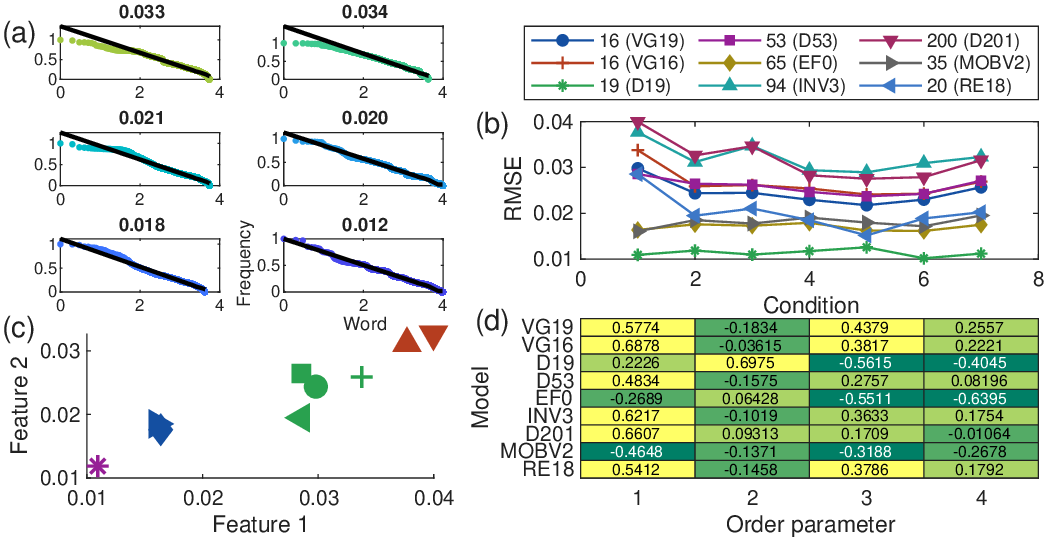}\caption{\textbf{Zipf’s law under different input conditions in Pre-CNNs.} The legend on the upper right defines different Pre-CNNs with the preceding number representing the number of convolutional layers. (a) Zipf's distributions under different RMSE levels. (b) Average RMSE of nine Pre-CNNs across seven conditions. (c) The performance of the Pre-CNNs in (b) clusters into four groups, suggesting shared feature extraction strategies despite their differences in architecture. (d) Visual order parameters: mean, variance, skewness, and kurtosis were averaged across images for each condition. Pearson correlations with model RMSEs reveal which visual statistics each group emphasizes.
}
\end{figure}

\subsection{In search of emergent statistical laws in images}
In this section, we examine the behavior of these statistical laws across nine Pre-CNNs by using the inputs from seven different datasets. For each pre-CNN, all reported statistics are computed from feature representations obtained by averaging over 16 samples per input condition. Since the coefficient of determination $R^2$ exceeds 0.92 for all models, we focus on the fitting quality as reflected by the root mean square error (RMSE). 

In Fig.~2(a), we showed the performance of Zipf's law across different RMSE values and summarized the results of nine Pre-CNNs across the databases presented in Fig.~2(b), which acted as conditions throughout this study. The Zipf's law performance of Pre-CNNs can be divided into four major groups. Using the RMSE across seven conditions as features, we performed K-means\cite{ahmad2007k} with K equal to four, and the results are presented in Fig.~2(c). This suggests that in terms of image feature analysis, the Pre-CNNs exhibit four distinct patterns in multi-target recognition consistency.

Based on this observation, we analyzed the relationship between RMSE variation and four statistical order parameters (OPs)\cite{reynaud2019second,emrith2010measuring} from visual neuroscience and statistics: mean, variance, skewness, and kurtosis, computed across conditions. Pearson correlation analysis was used to examine the association between RMSE changes and the average OP trends within the images of each condition. The results are shown in Fig.~2(d). Specifically, the first group D201 and INV3 shows RMSE variations primarily related to the mean; the second group VG16 and VG19 tends to be associated with mean and skewness. Although D53 and RE18 are not significant compared with the first two, these two OPs are higher than the other two. The third group EF0 and MOBV2 shows negative correlations for OPs other than the second one, and the fourth group D19 emphasizes features in OP2.



\begin{figure}
\centering
\includegraphics[width=13cm]{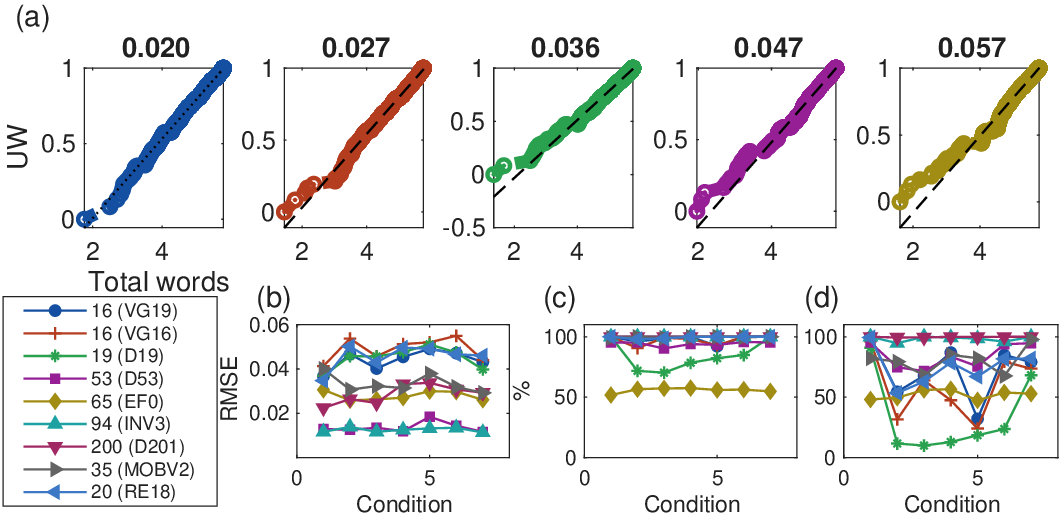}\caption{\textbf{Heaps’ law under different conditions.}
(a) Distributions of Heaps’ law under different RMSE thresholds.
(b) Performance of Heaps’ law following the original front-to-back input order.
(c) Proportion of cases with RMSE $< 0.02$ across 1,000 random permutations of feature map order.
(d) Same as (c) except RMSE $< 0.01$.
}
\end{figure}  

For Heaps' law, we similarly show the performance across different RMSE values in Fig.~3(a). Fig.~3(b) presents the average results of nine Pre-CNNs across seven conditions, which can be grouped into three major patterns. Unlike text, where the statistical properties of Heaps' law remain robust even after shuffling, the “words” in images depend on the sequential order of feature extraction and carry visual meaning, reflecting the perceptual relationships of foreground, middle, and background. Therefore, we cannot separate them as in traditional text. Here, we perform order shuffling by rearranging the order of feature maps. Figs.~3(c, d) show the proportion of RMSE values below 0.02 and 0.01, respectively, across 1000 iterations of feature map order permutations. Most groups maintain more than 50\% of cases with RMSE below 0.02, but at the 0.01 threshold, differences among Pre-CNNs become more pronounced.

\begin{figure}
\centering
\includegraphics[width=13cm]{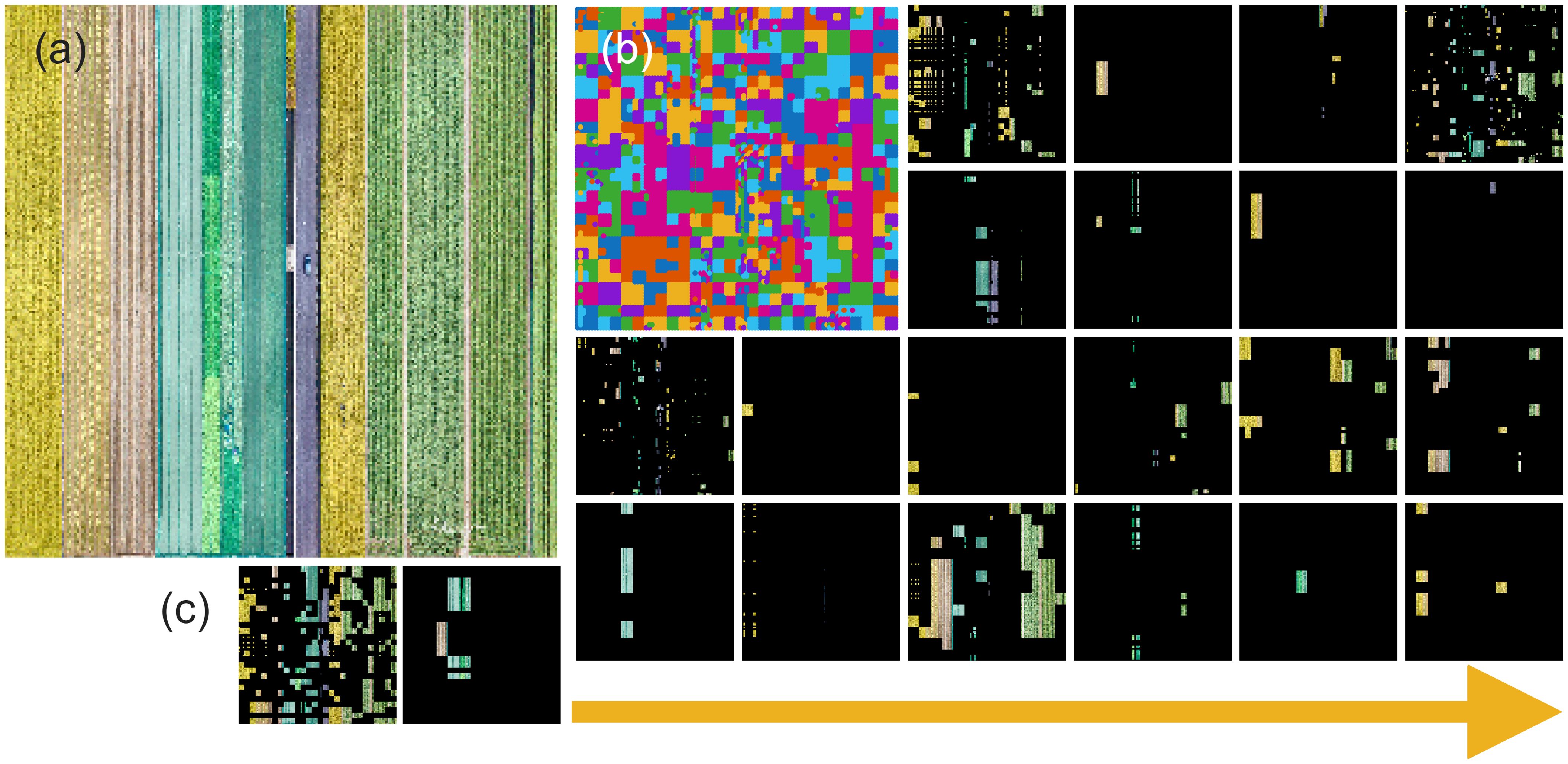}\caption{\textbf{Word-Position correlation  in ResNet-18.} 
(a) Original landscape image of Taiwan was authorized by Wei-Hsiung Huang (foto WH). 
(b) Pearson correlation is used to compute the relationship between the positions of word and each feature map activation, followed by segmentation with a 0.9 threshold. Correlated pixels primarily form small regions, reflecting the Zipf’s law that small regions constitute the main semantic components of the image. 
(c) To visualize the segmented correlated regions, four statistical order parameters are computed for each RGB channel, yielding 12 features per region. From the initial 4,800 feature maps, salient regions are selected and aggregated into 72 features, which are then clustered into 22 groups. The segmentation map shows the correspondence between these clusters and the original pixels from (b). 
}
\end{figure}  

To clarify the correspondence between Heaps' law and Zipf's law within the Pre-CNNs black box, we input a landscape photograph of Taiwan, shown in Fig.~4(a), into the RE-18 Pre-CNN for analysis. Following the Heaps' law approach, each feature map was resized to 112×112, and the sequence of feature maps was treated as a temporal order. Pearson correlation was computed between individual pixels, and correlations above 0.9 were used for image segmentation, as shown in Fig.~4(b). The resulting “words” appear in small localized regions, and due to upsampling, their positions correspond to actual locations in the image. This indicates that the emergence of Zipf's and Heaps' law patterns arises naturally from the contextual processing of image features within Pre-CNNs. We then performed K-means clustering with K equal to 22 on the RGB order parameters of 72 segmented regions from Fig.~4(b), and the results are shown in Fig.~4(c). For details, please refer to Sec. 5.4.

\begin{figure}
\centering
\includegraphics[width=13cm]{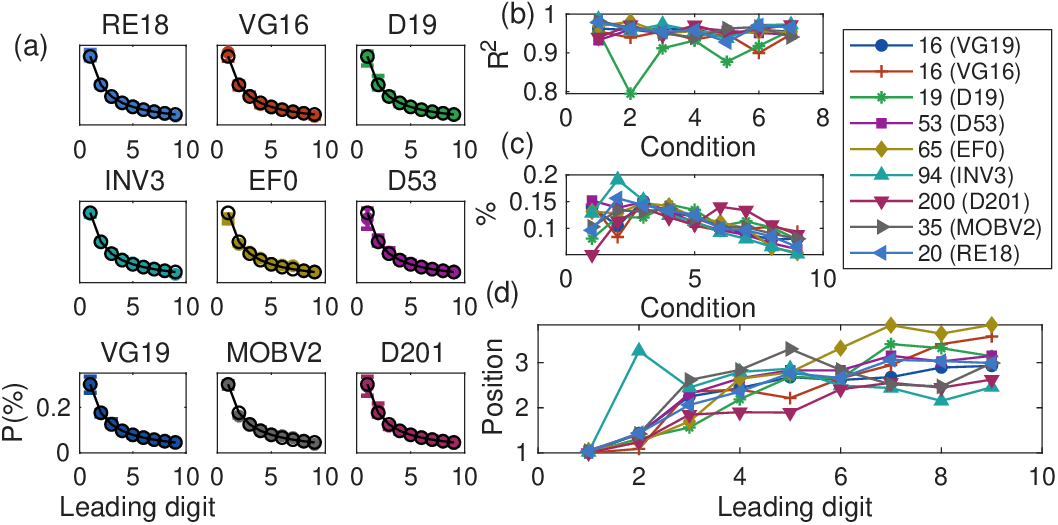}\caption{\textbf{Performance of Benford’s Law in Pre-CNNs.}(a) R-squared values of all Pre-CNN models under nine experimental conditions.
(b) Layer-wise proportion of the nine leading digits, averaged over 144 image inputs across nine conditions.
(c) Average layer positions of the leading digits based on the same setting as (b).
(d) These positions are further grouped into early, middle, and late stages using four layer partitions.}
\end{figure}

\begin{figure}
\centering
\includegraphics[width=13cm]{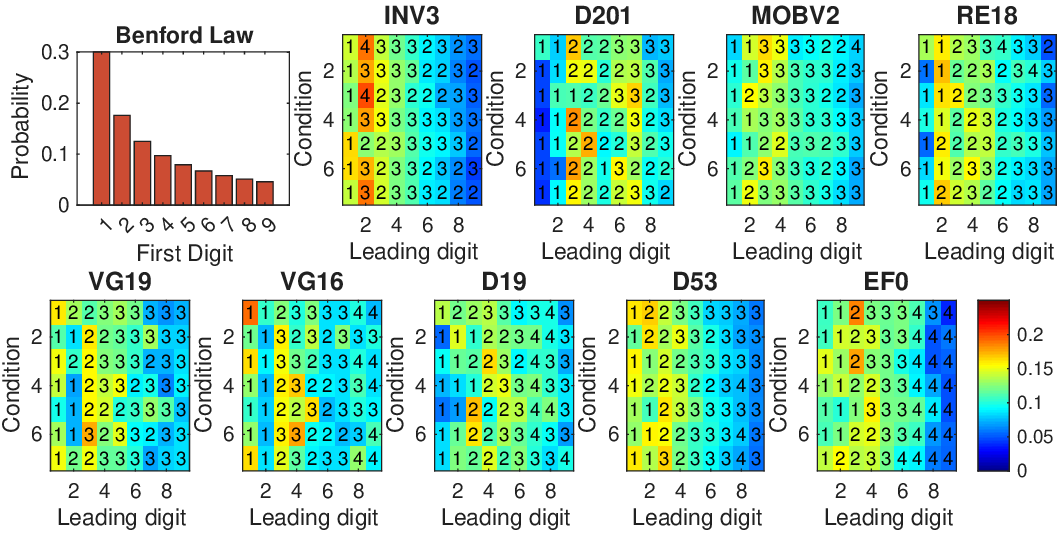}\caption{\textbf{Average layer positions of leading digits in Benford’s law.}
The leading digits associated with Benford’s law are analyzed across nine models using inputs from seven image datasets. For each model–dataset combination, the results are averaged over 16 samples. Convolutional Layer indices from 1 to 4 are further grouped to represent early, middle, and late stages of the network.
}
\end{figure}  

\begin{figure}
\centering
\includegraphics[width=13cm]{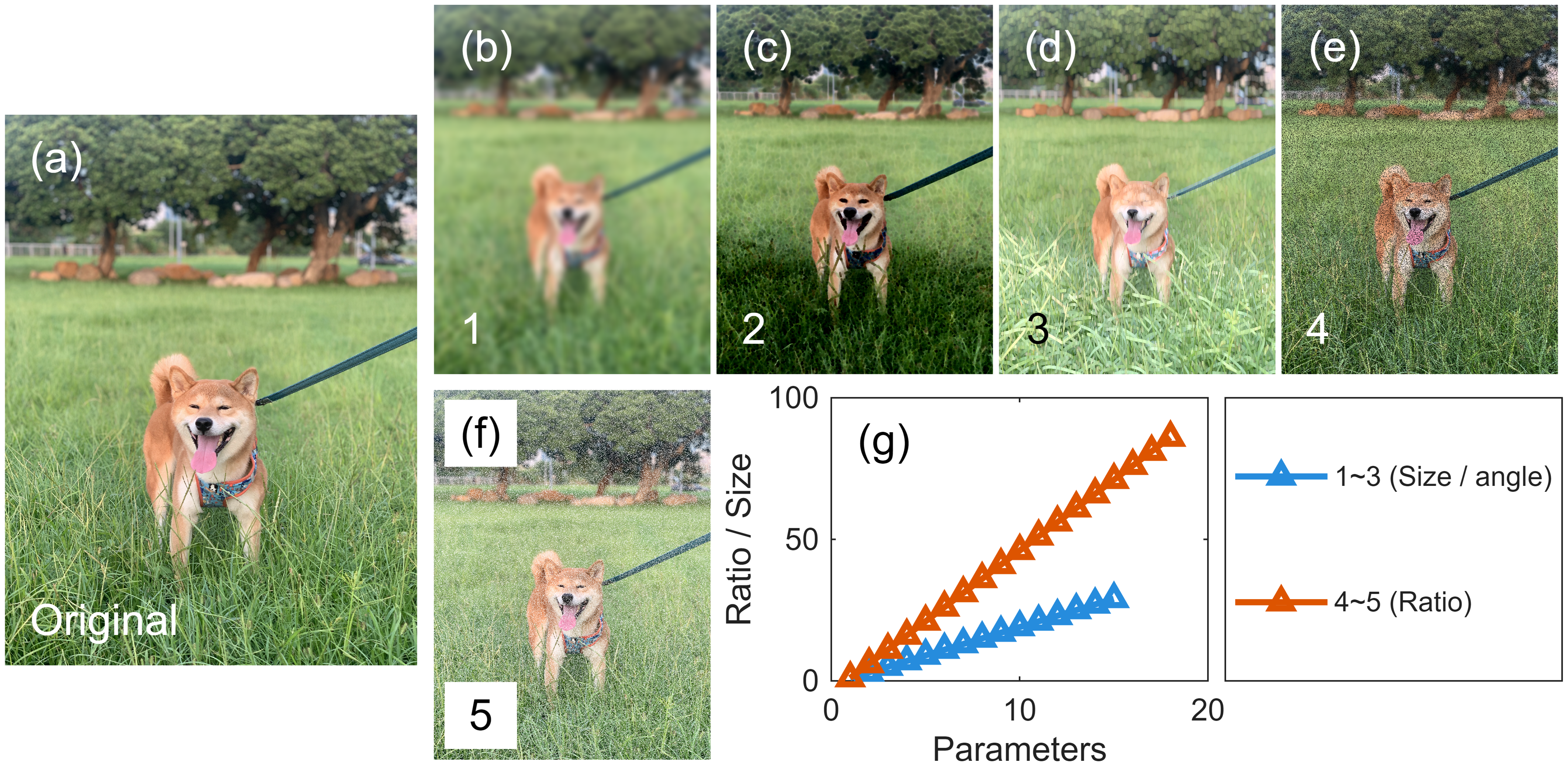}\caption{\textbf{Five attack methods used for robustness evaluation} The five methods correspond to (a)$\sim$(f). (a) Original image; (b) Gaussian blur; (c) erosion; (d) dilation; (e) random black pixel noise; (f) random white pixel noise; (g) parameters used in attacks 1–3 and 4–5.}
\end{figure}  

\begin{figure}
\centering
\includegraphics[width=13cm]{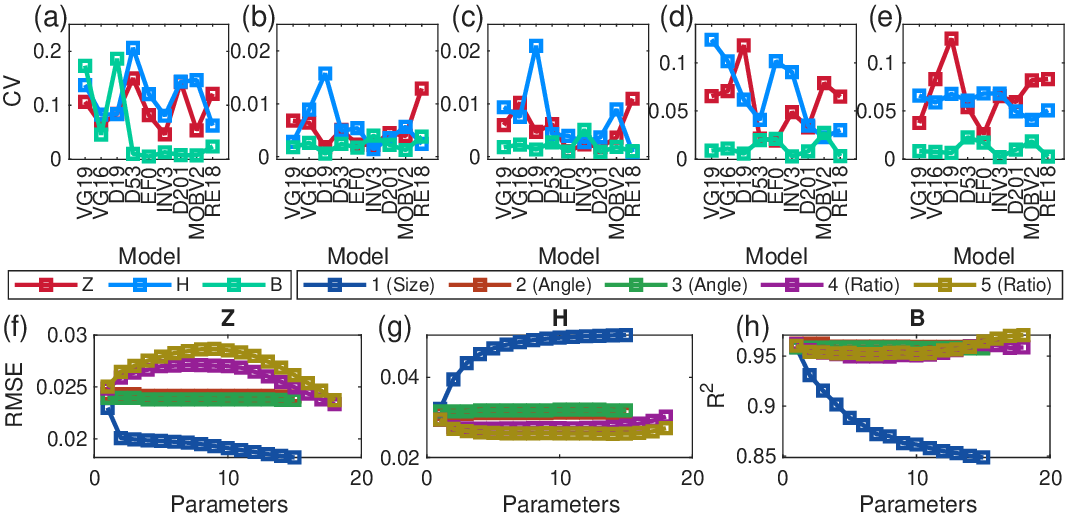}\caption{\textbf{Robustness under five image attacks.} (a$\sim$e) correspond to the five methods in Fig.~7(b$\sim$f). We use the Coefficient of Variation (CV) to evaluate whether the RMSE remains stable across attack sizes and proportions. Low CV indicates consistent variation across the group, while high CV indicates increased variability between instances. (f$\sim$h) show the performance of the three statistical laws under different parameters, averaged across all conditions and Pre-CNNs.}
\end{figure}

Benford's law, like the other power-law distributions, has been observed in many domains such as finance\cite{miller2015benford} and has recently been found to emerge in textual systems\cite{golbeck2023benford}, making it a target of our analysis. We adopted a strategy of combining adjacent convolutional layers into nine major groups and obtained the results by selecting the distribution that minimized the error while following the Benford's law pattern from higher to lower proportions, as detailed in Sec. 5.3. The average results across all samples for the nine Pre-CNNs are shown in Fig.~5(a).

We analyzed all Pre-CNNs across seven conditions and found that most R-squared values exceeded 0.9, indicating stable performance, as shown in Fig.~5(b). Regarding the algorithm for combining convolutional layers, we analyzed the distribution of layer counts across the nine leading digits in Fig.~5(c). Most layers were concentrated in the second leading digit, with counts gradually decreasing for higher digits. By performing a quartile analysis of the layer positions and averaging the results, we obtained Fig.~5(d), which demonstrates that the nine leading digits follow the sequential layers, giving rise to the emergence of Benford's law. The actual average layer positions under different conditions are shown in Fig.~6. With the exception of INV3, the patterns of Benford's law for the other Pre-CNNs follow the order of the leading digits consistently with that of layer combinations.


\begin{figure}
\centering
\includegraphics[width=13cm]{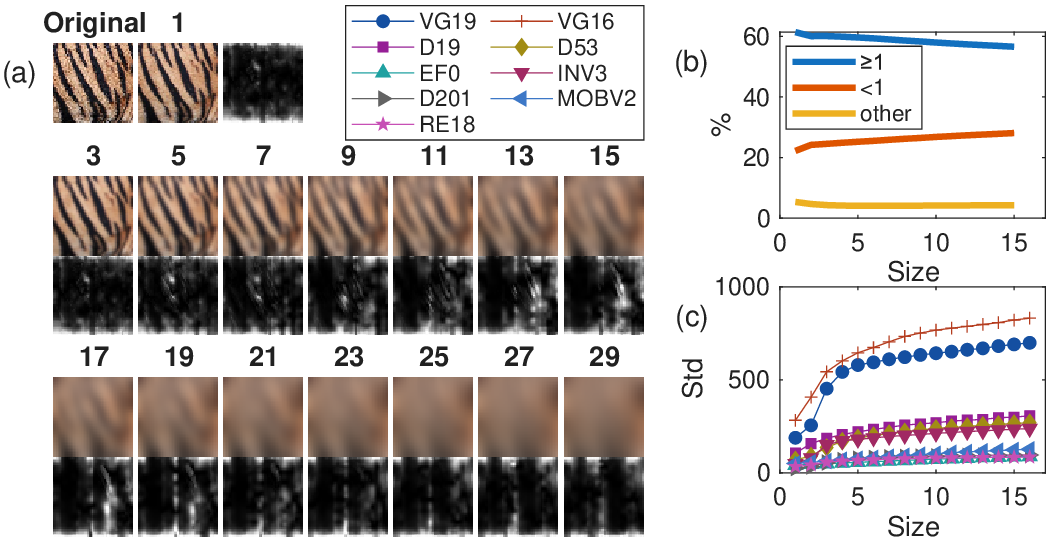}\caption{\textbf{Effects of Gaussian blur on the size of word bank in feature maps.} 
(a) Using a single image from Condition 2, we demonstrate the differences in the original image under Gaussian blur. The Arabic numbers indicate the kernel size of Gaussian blur, while the accompanying images show the normalized (0--1) average locations of words in each feature map, obtained by marking each feature map with 1 and averaging across all maps. The regions where words emerge are found to increase gradually. 
(b) Average word ratio of each feature map across all models and conditions are calculated to compare the original image with and without Gaussian blur. The ``other'' category represents VG19's feature maps without words. It is evident that the word count decreases continuously after the application of Gaussian blur. 
(c) The standard deviation of emergent word counts across layers, however, gradually increases.}

\end{figure}

\begin{figure}
\centering
\includegraphics[width=13cm]{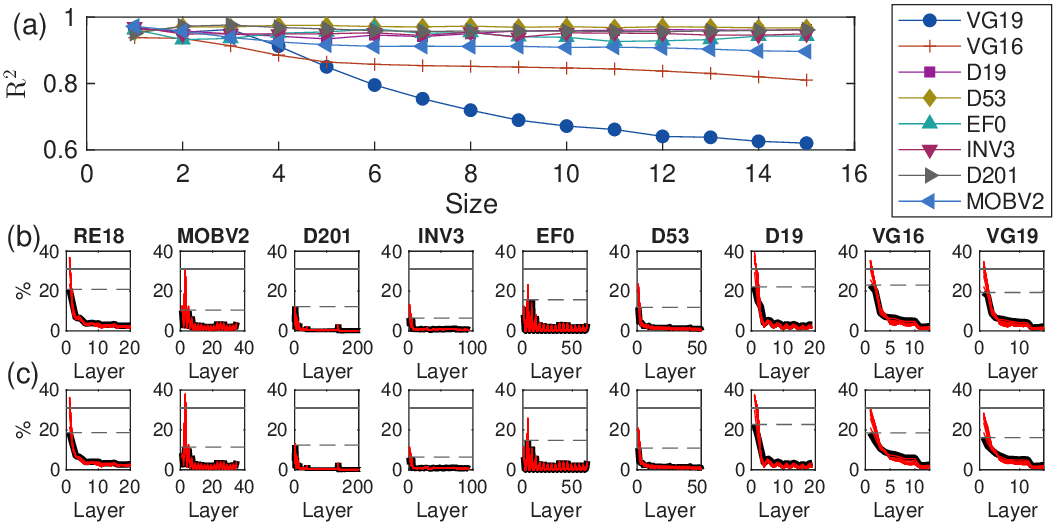}\caption{\textbf{Robustness of Benford's law under Gaussian blur attack across different Pre-CNNs.} 
(a) Performance of Benford's Law under Gaussian blur for different Pre-CNNs is measured by R-squared values. 
(b, c) Word count distributions across layers for a single image from Condition 4 \& 10 under different Pre-CNNs. The value for the original image are indicated by the black line to compare with the performance under attacks with various kernel sizes. The solid black line marks the 31\% position of the first leading ratio, and the dashed line indicates the layer with the highest word count in the original image. It is observed that, except for D201, INV3, EF0, and D53, all other Pre-CNNs exceed 31\% word count in a single layer under attack, indicating that even before combining word counts across layers, the network has already lost the possibility of fully fitting Benford's Law.}

\end{figure}  

\subsection{Test the robustness of these statistical laws} 
To evaluate the robustness of these three statistical laws in Pre-CNNs under different types of image perturbations, we applied five attack types with the corresponding results in Fig. 7(a$\sim$f). In previous studies, adversarial examples and image distortions\cite{gilmer2019adversarial,hendrycks2018benchmarking} have been commonly used to assess how deep learning models respond to corrupted inputs, providing insights into model sensitivity and the stability of feature extraction. Following this approach, we intentionally corrupted input samples to examine whether the three statistical laws remain stable under different perturbation conditions. The five attack types are as follows: Gaussian blur, erosion, dilation\cite{soille2004erosion}, and additive black-and-white noises. The kernel sizes and ratios for these attacks are shown in Fig. 7(g).

To quantify the stability of RMSE under different attack sizes and ratios, we calculated the Coefficient of Variation (CV), defined as the ratio of the standard deviation to the mean of RMSE for a given condition. Low CV indicates consistent RMSE across samples, reflecting relative stability in feature extraction, while high CV indicates increased variability between samples, suggesting that perturbations disrupt the statistical regularities. The results are shown in Fig. 8(a$\sim$ e), 
where erosion and dilation generally produce lower CV than the other attacks, and Benford's law tends to remain more stable than Zipf's and Heaps' laws across most conditions. In Fig. 8(f), taking Zipf’s law as an example, its stability may be primarily associated with low-frequency features. Under additive black-and-white noise, when the noise ratio is low, high-frequency regions increase, leading to higher RMSE; however, when the noise ratio exceeds approximately 50\%, large-scale merging of black and white regions occurs, promoting the formation of low-frequency areas and resulting in a decrease in RMSE. In Figs. 8(g, h), we observe that Gaussian blur actually reduces the fit of Heaps’ and Benford’s laws.


 
To unlock the ``black box'' of its effect,  we varied the strength of Gaussian blur and analyzed the average emergence locations of feature-map words, combined with upsampling and downsampling to a fixed resolution of $112 \times 112$. The results in Fig.~9(a) reveal that Gaussian blur not only alters the spatial distribution of emergent words for texture inputs, but also induces pronounced birth--death dynamics of words within individual feature maps.
Next, the results were averaged over all Pre-CNNs and experimental conditions in Fig.~9(b) which showed that the proportion of feature-map words generated by the attack exceeding those of the original images gradually decreases as the Gaussian kernel size increases. 
Furthermore, Fig.~9(c) indicates that the disparity in the number of generated words across different feature maps becomes progressively amplified with increasing blur strength - a trend consistently observed across all Pre-CNNs. 
This growing imbalance among feature maps directly impacts the manifestation of Heap’s law. Because Total Words and UW are accumulated at the feature-map level, 
the increasing variance in word production disrupts the stable power-law growth 
with a fixed exponent that Heaps' law would otherwise predict.

The degradation of Benford’s law in Pre-CNNs with increasing Gaussian blur kernel size is illustrated in Fig.~10(a). To further investigate this effect, we applied two different input images to nine Pre-CNNs in Figs.~10(b, c). Notably, five models exhibit at least one convolutional layer whose word-count proportion approaches or even exceeds the maximum probability of 31\% prescribed by Benford’s law after being attacked by the Gaussian blur. 
As a consequence, these models are driven into a regime in which a valid Benford-law fitting is theoretically impossible at the affected layer, even prior to any fitting procedure. This observation indicates that Gaussian blur induces structural deviations in the numerical statistics of feature-map activations that is irreconcilable with Benford’s law through the parameter optimization alone.



\section{Discussions}


How symbolic and written systems emerge from perceptual grounding is a central question shared across artificial intelligence, linguistics, semiotics, and psychology. In this work, we seek to address this question from the perspective of visual representation learning. Our work suggests that the emergence of Zipf’s, Heaps’, and Benford’s laws in Pre-CNNs aligns with the notion of articulation\cite{albrecht1999andre,chandler1994semiotics,barthes1977elements}: for an image to function like a linguistic system, it must be decomposable into perceivable and structured units\cite{liu2015hmax}, analogous to morphemes in written language. This supports the idea that statistical regularities observed in natural language can spontaneously arise from hierarchical feature representations in deep visual networks.

In Sec. 2.1, we demonstrate that these three statistical laws can emerge spontaneously in Pre-CNNs without any explicit symbolic background. This differs from previous studies, where solutions for symbol grounding in artificial intelligence or connectionist systems typically assume mixed-symbolic architectures\cite{harnad1990symbol}, which raises the question of whether the “first cause” of symbols must itself be symbolic. Our findings show that the internal information processing of connectionist networks alone is sufficient to generate the structural characteristics of symbols, addressing the first of three research objectives laid out in the ending paragraph of Introduction.

In Sec. 2.2, we not only examined the fitting performance of Zipf’s law in Fig. 2, but also identified that the information-processing patterns of nine common Pre-CNNs can be grouped into four major contexts, which correlate with the statistical OP of images. This provides a novel approach for interpretable deep learning that bridges image recognition and neuroscience insights, distinct from traditional feature-sampling-based interpretability methods. In Fig. 3, we analyze the characteristics of Heaps’ law and investigate the robustness of these statistical patterns when the order of feature maps is altered. Additionally, by retaining highly activated pixels above 90\% grayscale in Fig. 4, we found that most segmented visual words consist of many small regions rather than a few large ones, reflecting the power-law properties inherent in both Zipf’s and Heaps’ laws. The results for Benford’s law are presented in Figs. 5 and 6, where we analyze how the integration of different layers and proportions across Pre-CNNs gives rise to statistical regularities. Taken together, across all conditions—whether related to surface textures or objects—the three statistical linguistic laws consistently emerge, corresponding to the second research objective.

In Sec. 2.3, we conducted a robustness analysis of Pre-CNNs under five types of image perturbations in Fig. 7 to address the third research objective. Figures 8(a–e) present the fluctuations in fitting performance of the three statistical laws across different attack types, where the CV of the RMSE over varying parameters is used as an indicator. Among the three laws, Benford’s exhibits the highest robustness, showing the smallest performance variation across perturbations. In contrast, Gaussian blur induces the greatest instability in the statistical behavior of Pre-CNNs, indicating that smoothing-based degradations most strongly disrupt the underlying feature representations.

In Figs. 8(f–h), we further decompose the effects of different attack parameters by averaging across Pre-CNN architectures and input image conditions. This analysis reveals that Gaussian blur has a particularly strong impact on Heaps’ and Benford’s laws, leading to the largest deviations in fitting performance. Interestingly, Gaussian blur turned out to stabilize Zipf’s law. We surmised that this was attributed to the increase in low-frequency components caused by blurring, which promotes more homogeneous activation distributions and strengthens the power-law relationship. Our speculation is supported by Fig. 8(f), where variations in the proportion of black and white noises produce a non-monotonic RMSE trend - first increasing and then decreasing — indicating a similar low-frequency-dominated stabilization mechanism for Zipf’s law.

To further elucidate the mechanisms underlying the instability when facing image perturbations, we presented a detailed analysis in Figs. 9 and 10. They showed that Gaussian blur substantially amplifies the variance in the number of emergent visual words across feature maps. This increased heterogeneity disrupts the cumulative growth process required by Heaps’ law: as feature maps are progressively aggregated, the  power-law scaling between the number of UW and the total word count becomes unstable, leading to significant deviations in the fitting performance.

For Benford’s law, Gaussian blur produces a different but related failure mode. In several Pre-CNNs, such as RE18\cite{he2016resnet}, MOBV2\cite{sandler2018mobilenetv2}, D19\cite{redmon2016darknet}, VG16, and VG19\cite{simonyan2014vgg}, the earliest convolutional layers already contain a proportion of visual words that approaches or even exceeds the theoretical maximum expected for the leading digit 31\%. As a result, even before the application of the layer-integration procedure, the first-digit distribution becomes saturated, preventing accurate adherence to Benford’s law after aggregation.

In contrast, models such as D53\cite{redmon2016darknet}, EF0\cite{tan2019efficientnet}, INV3\cite{szegedy2016rethinking}, and D201\cite{huang2017densenet} are more robust under the Gaussian-blurred attacks. These architectures incorporate distinctive design features, including residual or dense connections, multi-branch convolutions, and optimized feature reuse mechanisms, which promote a more even distribution of feature extraction across layers. Consequently, no single convolutional layer dominates the first-digit statistics, allowing these models to maintain stable compliance with Benford’s law despite substantial image degradation. By comparison, single-path architectures tend to concentrate feature extraction within specific layers, rendering them more susceptible to perturbation-induced deviations from Benford’s law.

We not only investigated the spontaneous emergence of linguistic statistical laws in Pre-CNNs, but also analyzed the internal information processing within deep learning models. In Figs. 2(b,c) and 3(b), despite differences in network design, the processing of input images across different Pre-CNNs exhibited convergent clustering patterns. Furthermore, Figs. 9 and 10 show that Gaussian blur significantly alters feature distributions and weakens clear contours and details in images, making Heaps’ law and Benford’s law particularly susceptible to disruption. This suggests that the formation of visual words, resembling written characters, relies on well-defined local contours and lines, which are prone to blurring when excessively smoothed. Consequently, features representing these visual words are weakened, destabilizing the statistical laws. The findings indicate that statistical laws are highly sensitive to image boundary features, a property that may provide useful guidance for the decomposition and feature analysis of known written characters—for example, in the classification of plastic signs, iconic signs\cite{groupe2015tratado}, and visual semiotics\cite{aiello2020visual}, where features can be mapped to the distributional properties of the three statistical laws.

In this study, we confirmed that deep learning models that are trained solely on visual inputs spontaneously exhibit the same statistical properties as in linguistics, i.e., the Zipf’s, Heaps’, and Benford’s laws. Importantly, these patterns  arise naturally from hierarchical processing of perceptual features,  without direct exposure to textual or linguistic data. This indicates that the internal representations of the models contain implicit symbolic organization, providing evidence that machines can generate quasi-symbolic units grounded in perception rather than in language itself. Notably, our models completely circumvent the “Chinese Room”\cite{searle2009chinese}, since they do not rely on prior symbolic system input, but instead allow statistical structures to emerge directly from perceptual data, contrasting the conventional approaches that require explicit symbolic input to construct concepts.

The symbol–world mapping problem has long been a central topic in artificial intelligence\cite{qiu2022emergent} and cognitive science\cite{harnad1990symbol,steels2008symbol}. Beyond the statistical properties of the symbolic system itself, related research has explored various grounding issues, such as visual grounding (mapping images to text)\cite{liu2025survey,xiao2024towards}, language grounding in robotics\cite{henry2024unsupervised}, vector grounding (embedding symbols in vector space)\cite{mollo2023vector}, and numerical grounding (conceptual grounding of numbers)\cite{leibovich2016symbol}. These studies emphasize direct correspondences between symbols and the perceptual or operational world, highlighting the crucial role of grounded representations in intelligent systems. However, we shifted the focus by investigating the emergence of linguistic statistical structure within purely image-based connectionist deep learning models, exploring internal statistical patterns of the symbolic system rather than direct symbol–world correspondences. This provides a complementary perspective, showing that even in the absence of explicit symbol–world mapping, deep learning models are still capable of retaining the statistical properties of language-like structures autonomously.


Within the conceptual framework of Steels in 2007\cite{steels2008symbol}, the representations identified here are best characterized as subsymbolic c-representations, which, despite lacking full m-symbol properties such as explicit convention, communicative intent, and negotiated meaning, exhibit systematic distributional regularities justifying their description as proto-symbolic. These constitute structured, reusable representational units that precede and constrain later symbolic assignment, rather than resulting from it. Building on Steels’ argument that symbol grounding occurs when symbols are linked to operational perceptual procedures, our results extend this view by demonstrating that symbol-like statistical structures can arise prior to explicit linguistic symbol assignment. In this setting, visual perception itself provides a grounded substrate from which proto-symbolic organization emerges spontaneously. Observed adherence to Zipf’s, Heaps’, and Benford’s laws indicates that the relationships between images, symbols, and language-like statistical properties are not arbitrary, but reflect deep regularities in perceptual feature distributions.

From an information-processing perspective, perceptual processing itself is inherently context-dependent. From the primary visual cortex (V1) to higher-level visual areas \cite{rolls2023multiple}, such as the inferotemporal (IT) cortex \cite{arcaro2021relationship}, the visual system exhibits a hierarchical organization spanning early, intermediate, and late stages of perception. Meaningful perception does not arise in isolation at any single level, but instead emerges through interactions across different levels, in which contextual relationships play a decisive role in shaping perceptual interpretation. Language, as a highly organized symbolic system, exhibits a closely analogous property: meaning is not intrinsic to isolated symbols, but is established through relational networks embedded within broader contextual structures \cite{groupe2015tratado,sadeghi2015you}.

As a model highly aligned with human visual feature processing, Pre-CNNs naturally embody these principles in their internal mechanisms, including hierarchical organization and bottom-up information flow \cite{eickenberg2017seeing}. These characteristics provide a crucial theoretical foundation for defining image-based ``words'' and for understanding how statistical linguistic laws emerge from visual representations. Accordingly, the definition of visual sign units cannot be based on features extracted from a single-layer feature map alone. Functionally, feature maps in Pre-CNNs are more analogous to morphemes in language, while only the salient and dominant features that emerge through hierarchical propagation and two-dimensional convolutional processing can serve as primary representative ``words.''

In this context, our modeling framework is consistent with the systematic theory of visual signs proposed by the Belgian semiotics research group Groupe $\mu$ \cite{groupe2015tratado}, which emphasizes that visual signs are not isolated objects but structures of signification established through relationships among elements. These elements, referred to as entities (entidades), can be hierarchically organized into units, sub-entities (subentidades), and supra-entities (supraentidades), forming a multi-level structural organization. This relational and hierarchical perspective also underlies the concept of plastic signs, in which meaning arises not from direct referential depiction of external objects, but from formal and structural relations themselves.


The hierarchical organization of visual sign units is thus constructed through contextual relations inherent to signification itself. In other words, hierarchies emerge through comparison, contrast, and relational interaction among units, rather than being predefined or imposed a priori. This view closely aligns with the principles of connectionist neural networks, particularly in their preprocessing stages, where information does not reside in individual nodes but emerges from patterns of relations embedded within a broader contextual structure. From this perspective, the spontaneous emergence of text-like or language-like structures in images may be understood as a direct consequence of perceptual and informational processing mechanisms that are fundamentally relational and context-driven.

Taken together, these findings suggest that, in our study, the symbol grounding problem is conceptualized as the emergence of statistically organized quasi-symbolic representations during the hierarchical and sequential propagation of image features, which is then characterized through the distributions of statistical linguistics.
In particular, Benford's law, which has previously been applied to detect fabricated numbers and AI-generated texts\cite{wang2025novel}, may serve as a useful tool for differentiating AI-generated images from real-world videos, thereby potentially contributing to the development of robust verification methods. The distinction and correspondence between writing systems, images, and marks will also be the focus of our future work.


\section{Conclusion}\label{sec13}
Our modeling approach is inspired by the hierarchical nature of visual information processing and the retinal imaging mechanism underlying human vision. Although it is currently impossible to directly observe dynamic, global information processing in the human brain using non-invasive methods due to limited spatial and temporal resolution, recent advances in Pre-CNNs have been shown to closely correspond to the transmission of critical visual features in the human brain and successfully applied in the brain--computer interface research. For images to function as a form of language, they must be organized hierarchically into structural units, from small to large, revealing interpretable relationships within the image.


Research in artificial intelligence and cognitive science has explored the reproducibility of the evolution from images to abstract symbols. Cognitive studies suggest that writing systems gradually evolved from pictorial signs to abstract symbols. For example, early humans depicted the sun using sketches approximating its natural form and, through repeated communication and interaction, gradually linked visual concepts to symbolic representations, eventually forming new shared symbol systems. Experiments such as Pictionary-style communication games simulate this process, demonstrating how iterative interaction can generate a symbol system, while balancing accuracy and efficiency. These studies highlight the environmental and interactive conditions necessary for forming human-like graphic symbol systems\cite{qiu2022emergent}.

Collectively, our findings suggest that the hierarchical feature representations in Pre-CNNs may, to some extent, recapitulate this evolutionary process. Just as humans transform perceptual sketches into abstract symbols through structured interaction, deep visual networks can organize low-level features into structured proto-symbolic units, giving rise to statistical regularities analogous to those found in natural languages. This supports the idea that images themselves can serve as a foundation for symbol generation and spontaneously form language-like structures.

\section*{Acknowledgement}
 We thank Wei-Hsiung Huang (foto WH) for providing two beautiful photographs of Taiwan for use in this academic work.
We are grateful to the financial support from the National Science and Technology Council in Taiwan under Grants No. 113-2112-M007-008 and 114-2112-M007-004.

\section{Methods}\label{sec11}
\subsection{Datasets}
All conditions were conducted using publicly available image datasets. Texture images were obtained from the Describable Textures Dataset (DTD) curated by JMExpert on Kaggle. Additional images were sourced from the CV-Assignment3-Images dataset by anasahmad25 and the Aquarium Dataset by Sharan Sajiv Menon. The Berkeley Segmentation Dataset and Benchmark (BSDS500), provided by the University of California, Berkeley, was used for natural image segmentation experiments. All datasets are distributed under open or permissive licenses, including the Apache License, Version 2.0, the Community Data License Agreement – Permissive, Version 1.0 (CDLA-Permissive-1.0), and the Creative Commons CC0 1.0 Universal public-domain dedication. All data were used in accordance with their respective licensing terms. Figures 1 and 4 | Taiwan landscape photograph, reproduced with authorization from Wei-Hsiung Huang (foto WH). Figure 7 is the photograph of Chi-chi Huang, taken by the first author’s spouse, Yu-Hsuan Kao, and reproduced with authorization from the pet’s owner, Xin-Ying Huang.

\subsection{Defining image words and the emergence of Zipf's and Heaps' laws}
Building on prior work in explainable deep learning, we identify the most prominent features in each convolutional feature map by selecting pixels with activation values exceeding 90\% of the maximum in accordance to the common strategy in feature visualization and saliency analysis \cite{selvaraju2017gradcam,zhou2016learning}. To evaluate the spatial consistency of these selected regions, we adopt Intersection over Union (IoU) \cite{van2019deep} as a reference metric. By focusing on the top-activated pixels, we estimate the frequency of visual ``words,'' observing the power-law distribution consistent with Zipf's law. Furthermore, sequentially counting the cumulative number and types of these words allows Heaps' law to naturally emerge. All analyses are conducted across seven open-source image databases using pre-trained CNN architectures (Pre-CNNs).
\subsection{Algorithm of emerging Benford's law}
In this approach, we focus on optimizing the distribution of first digits by merging adjacent layers until the distribution consists of exactly 9 groups. The algorithm starts by normalizing the input distribution so that the sum of all values equals 1, converting it into a valid probability distribution. Initially, each element of the distribution is treated as an individual group. We then proceed by merging adjacent groups iteratively. In each iteration, we calculate the fit of the newly merged distribution to the target first-digit distribution, i.e., the distribution representing the first digits 1 through 9. The goal is to minimize the difference between the merged distribution and the expected first-digit pattern. The quality of the merging process is evaluated using the \( R^2 \) value. The algorithm merges the two adjacent groups that provide the best \( R^2 \) value after their combination, ensuring the best fit to the target first-digit distribution at each step. The algorithm continues the merging process until exactly 9 groups remain, with each corresponding to one of the first digits from 1 to 9. This method ensures that the input distribution is transformed into one with 9 groups, each of which represents one of the first digits, optimizing the fit to the expected first-digit distribution.

\subsection{Heaps' law and image segmentation}
Through Heaps' law, we can determine the order of each kernel. We then upsample the feature map to the original image size, where the brightness distribution of each pixel will change according to the order. Using this, we can calculate the Pearson correlation to assess the correlation and perform segmentation based on the connectivity properties of the graph. Finally, we use statistical features from the RGB channels—mean, variance, skewness, and kurtosis—to perform k-means clustering for the segmentation results. For details, please refer to Section 5.5.

\subsection{Robustness evaluation and Parameter Settings}
To systematically probe the robustness of three statistical linguistic laws within the internal representations of Pre-CNNs, five types of image perturbations were applied with explicitly controlled parameter ranges. For each perturbation, the severity level was gradually increased according to predefined step sizes, ensuring consistent and reproducible distortion strength across all models and datasets. All perturbations were applied to randomly cropped image patches matching the input resolution of each network.

Gaussian blur was used to simulate progressive degradation of fine-grained visual details. Each cropped image was convolved with an isotropic Gaussian filter, where the standard deviation $\sigma$ controlled the blur strength. The parameter $\sigma$ was varied from 1 to 30, with a step size of 2, progressively suppressing high-frequency texture information while largely preserving global luminance structure. Morphological erosion was performed using a linear structuring element with a fixed length of 11 pixels, where the orientation angle of the structuring element was swept from $1^\circ$ to $30^\circ$ in increments of $2^\circ$. This operation progressively removes bright regions and thins object boundaries, thereby disrupting local spatial continuity and fine structural details. Morphological dilation employed the same linear structuring element configuration and orientation range as erosion, expanding bright structures and thickening edges, often causing nearby features to merge.

To introduce stochastic pixel-level noise, random black pixel destruction was applied by randomly selecting a fixed percentage of pixels and setting their intensities to zero across all color channels. The destruction ratio ranged from 1\% to 90\% of the total number of pixels, with increments of 5\%, producing spatially uncorrelated impulsive noise. In a complementary manner, random white pixel destruction set a fixed percentage of randomly selected pixels to maximum intensity across all channels, with the same percentage range and step size, introducing salt-like noise and high-intensity outliers. Together, these perturbations span linear filtering, non-linear morphological transformations, and stochastic pixel-level noise, enabling a controlled assessment of the robustness of emergent statistical linguistic regularities in Pre-CNN representations under diverse image distortions.





\bibliography{sn-bibliography}

\end{document}